\documentclass[nonacm, sigconf]{acmart}

\usepackage{algpseudocode, algorithm}
\makeatletter
\renewcommand{\ALG@beginalgorithmic}{\small}
\makeatother
\usepackage{graphicx}
\usepackage{textcomp}
\usepackage{xcolor}
\usepackage{subcaption}
\usepackage{multirow}
\usepackage{array, booktabs, longtable}

\usepackage{tikz}
\usepackage{tikz-qtree,tikz-qtree-compat}
\usetikzlibrary{shapes,arrows,positioning,calc}

\tikzset{decision/.style={diamond, draw, fill=blue!20, text width=4.5em, text badly centered, inner sep=0pt}}
\tikzset{block/.style={rectangle, draw, fill=blue!20, text width=5em, text centered, rounded corners,
 minimum width=3.5cm}}
\tikzset{line/.style={draw, -latex}}

\AtBeginDocument{%
  \providecommand\BibTeX{{%
    \normalfont B\kern-0.5em{\scshape i\kern-0.25em b}\kern-0.8em\TeX}}}

\copyrightyear{2022}
\acmYear{2022}
\setcopyright{rightsretained}
\acmConference[DAC '22]{Proceedings of the 59th ACM/IEEE Design Automation Conference (DAC)}{July 10--14, 2022}{San Francisco, CA, USA}
\acmBooktitle{Proceedings of the 59th ACM/IEEE Design Automation Conference (DAC) (DAC '22), July 10--14, 2022, San Francisco, CA, USA}
\acmDOI{10.1145/3489517.3530620}
\acmISBN{978-1-4503-9142-9/22/07}

\acmConference[Conference acronym 'XX]{Make sure to enter the correct
  conference title from your rights confirmation emai}{June 03--05,
  2018}{Woodstock, NY}
\begin{document}

\title{On Robustness and Generalization of ML-Based Congestion Predictors to Valid and Imperceptible Perturbations}

\author{Chester Holtz}
\authornote{Corresponding author}
\email{chholtz@eng.ucsd.edu}
\affiliation{%
\institution{University of California San Diego}
\country{}
}

\author{Yucheng Wang}
\email{yuw132@ucsd.edu}
\affiliation{%
\institution{University of California San Diego}
\country{}
 }
 
\author{Chung-Kuan Cheng}
\email{ckcheng@eng.ucsd.edu}
\affiliation{%
\institution{University of California San Diego}
\country{}
 }

\author{Bill Lin}
\email{billlin@eng.ucsd.edu}
\affiliation{%
\institution{University of California San Diego}
\country{}
 }

\renewcommand{\shortauthors}{Pengwen Chen et al.}

\begin{abstract}
There is substantial interest in the use of machine learning (ML)-based techniques throughout the electronic computer-aided design (CAD) flow, particularly methods based on deep learning. However, while deep learning methods have achieved state-of-the-art performance in several applications (e.g. image classification), recent work has demonstrated that neural networks are generally vulnerable to small, carefully chosen perturbations of their input (e.g. a single pixel change in an image). In this work, we investigate robustness in the context of ML-based 
EDA tools\textemdash particularly for congestion prediction. As far as we are aware, we are the first to explore this concept in the context of ML-based EDA.

We first describe a novel notion of imperceptibility designed specifically for VLSI layout problems defined on netlists and cell placements. Our definition of imperceptibility is characterized by a guarantee that a perturbation to a layout will not alter its global routing. We then demonstrate that state-of-the-art CNN and GNN-based congestion models exhibit brittleness to imperceptible perturbations. Namely, we show that when a small number of cells (e.g. 1\%\textemdash 5\% of cells) have their positions shifted \emph{such that a measure of global congestion is guaranteed to remain unaffected} (e.g. 1\% of the design adversarially shifted by 0.001\% of the layout space results in a predicted decrease in congestion of up to 90\%, while no change in congestion is implied by the perturbation). In other words, the quality of a predictor can be made arbitrarily poor (i.e. can be made to predict that a design is ``congestion-free'') for an arbitrary input layout. Next, we describe a simple technique to train predictors that improves robustness to these perturbations. Our work indicates that CAD engineers should be cautious when integrating neural network-based mechanisms in 
EDA flows to ensure robust and high-quality results. 
\end{abstract}
\maketitle

\section{Introduction}
\label{sec:introduction}

Electronic design automation (EDA) flows involve  significant optimization and verification challenges that continue to scale as the complexity of designs increases. There is substantial interest in using machine learning techniques for solving electronic computer-aided design (CAD) problems ranging from logic synthesis to physical design and design for manufacturability (DFM)~\cite{darpa2018}. 
Prior work has demonstrated that deep learning-enhanced design flows are faster and more scalable, particularly when integrated to augment the time-consuming stages of layout~\cite{mavirec, congestion, routenet}, design space exploration~\cite{greathouse2018mlhetero}, logic optimization~\cite{yu2018synthesis} and lithographic analysis~\cite{yu2012hotspot}.
%

However, although neural networks have been extremely successful in the aforementioned EDA tasks, recent work~\citep{szegedy2014intriguing, goodfellowadv} has demonstrated that 
image classifiers can be fooled by small, carefully chosen perturbations of their input. Notably, \citet{SuSinglePixel17} demonstrated that neural network classifiers which can correctly classify ``clean'' images may be vulnerable to \textit{targeted attacks}, e.g., misclassify those same images when only a single pixel is changed.

The question that we aim to explore in this work is the following:
\begin{quotation}\textbf{\emph{To what degree are neural network-based congestion predictors vulnerable to small, but valid, changes in layout input?}}
\end{quotation}
As the application of machine learning to production EDA tasks becomes more widespread, understanding and addressing this question will become increasingly critical. In this work, we provide evidence that supports an affirmative answer:
\begin{quotation}\textbf{\emph{
Congestion predictors erroneously predict large changes to routing congestion with respect to changes to the layout that do not change the global routing.
}}\end{quotation}
Specifically, we investigate a novel notion of validity and design two efficient methods for finding perturbations that demonstrate brittleness of recently proposed congestion predictors. Furthermore, we describe one potential approach to address the highlighted issues and demonstrate that modifying the training procedure to promote robustness is one promising direction to address brittleness to imperceptible changes. Although we focus on congestion prediction, our work generalizes to arbitrary predictive models integrated in EDA pipelines. More generally, our work motivates the need for careful evaluation of the generalization of ML-based EDA tools\textemdash in excess of typical performance metrics reported on a train-test split.

\subsection{Contributions}

The primary contribution of this work is to demonstrate that modern deep learning-based EDA tools\textemdash specifically congestion predictors\textemdash are vulnerable to valid perturbations to their inputs, i.e. may exhibit poor generalization to perturbations of the cell layout. 

Inspired by the perspective of \emph{adversarial perturbations}, given an input design layout, we characterize \emph{small} perturbations as (1.) perturbations that result in the adjustment of relatively few cell positions and (2.) perturbations that maintain the global congestion structure (Sec.~\ref{sec:method}). We describe a numerical algorithm to efficiently search the \emph{feasible adversarial neighborhood} of an input to find small perturbations that maintain validity of the input design while drastically reducing the efficacy of ML-based EDA tool predictions (Sec.~\ref{sec:method}). We describe two variants of the proposed method: while both rely on knowledge of the predictor weights\textemdash known as a white-box perturbation model\textemdash one method requires knowledge of the underlying congestion structure, while the second method does not necessitate such information. We then demonstrate (1.) that congestion predictors are vulnerable to both models and (2.) that \emph{adversarial training} significantly improves robustness with only a modest performance trade-off. We emphasize that while we frame our discussion in the context of robustness to perturbations, our findings motivate a broader need to study implications of poor generalization when integrating ML-based tools into design flows. 

In summary, our contributions include the following:
\begin{enumerate}
    \item A novel formulation of the \emph{feasible neighborhood} of an input design\textemdash i.e. given a layout, what small perturbations maintain the relevant measures of congestion?
    \item Efficient supervised and unsupervised algorithms for computationally searching the neighborhood of a layout.
    \item Exploration of adversarial training as a way to induce robustness and improve generalization.
    \item Under a previously defined characterization of predictive quality for congestion tasks~\cite{fpgacongestion}, we show that the benchmark layouts we evaluated can be perturbed such that the congestion predictions on the perturbed layouts are poor.
\end{enumerate}

\section{Preliminaries and Related Work}
\label{sec:preliminaries}

In this section, we provide an overview of ML-based EDA methods and adversarially robust prediction.
\begin{figure}[!ht]
\fbox{\begin{minipage}[t]{0.45\textwidth}
    Number of components \hfill $n \in \mathbb{R}_+$ \\
    Placement coordinates \hfill $x, y \in \mathbb{R}^n$, $X = [x:y]\in \mathbb{R}^{n\times 2}$ \\
    Placement perturbation \hfill $\delta_x, \delta_y \in \mathbb{R}^n$, $\Delta = [\delta_x:\delta_y]\in \mathbb{R}^{n\times 2}$ \\
    Neural network parameters \hfill  $\theta$ \\
    Early global routing bins \hfill $W \times H$ \\
    Feature map \hfill  $M \in \mathbb{R}^{W\times H}$ \\
    Predicted congestion map \hfill  $f_\theta(M)$
\end{minipage}}
\caption{Notation}
\label{tab:notation}
\end{figure}
Let $x, y\in \mathbb{R}^{n}$ be vectors corresponding to the coordinates of $n$ components such that the $i$-th component has coordinates encoded in the $i$-th row of $X:=[x:y]$; $[x:y]_i$. We aim to find perturbations to the layout so that the resulting layout satisfies certain constraints (i.e. remains in the neighborhood of the original layout with respect to global routing).

\subsection{Global routing}
The VLSI routing problem is usually solved in two steps: (1.) global routing and (2.) detailed routing. The principle aim of the global routing step is to generate a routing solution on a discretization of the layout space, represented as a grid graph and provide a preferred routing region (i.e. a route guide) for the detailed router. 

A typical multi-commodity flow formulation of global routing partitions the routing space into regular rectangles (G-Cells) and generates a grid graph $G=(V, E)$ in which each vertex $v \in V$ represents a G-Cell and each edge $e \in E$ represents the connection between adjacent G-Cells. The capacity of an edge represents the maximum number of wires that can go through the edge and the variable assignment of the edge corresponds the number of wires that are currently using the edge, while the overflow is denoted by the number of wires that exceeds the capacity.
For each net, the routing problem is to find a path that connects all the pins of a net in the given grid graph while avoiding overflow on the edges. In other words, the global router maximizes a measure of routability with respect to the detailed router while satisfying certain constraints to manage design rule violations (DRVs), pin accessibility, and irregular module geometries. An important concept is the construction of the graph $G$. The graph, global routing solution, and associated congestion metrics remain consistent as long as individual cells remain within their G-Cells, regardless of their precise positions within each G-Cell.

\subsection{RUDY}

Rectangular Uniform wire Density (RUDY)~\cite{rudy} is a method to estimates the wirelength density by uniformly spreading the wire volume of nets into its bounding box. It is very commonly used as a feature to indicate the relative congestion of a region. The RUDY map of a net $e$ represents the average wirelength per unit area in the bounding box of the net:
$
\mu^{(e)} (\frac{1}{x_\text{max} - x_\text{min}} + \frac{1}{y_\text{max} - y_\text{min}})
$
where the net-map $\mu^{(e)}\in\mathbb{R}^{W\times H}$ is
$$
\mu^{(e)}_{xy} = \begin{cases}
1 & x_\text{min} \leq x \leq x_\text{max} \text{ and } y_\text{min} \leq y \leq y_\text{max}  \\
0 & \text{otherwise}
\end{cases}
$$
$x_{\min}$, and $x_{\max}$ correspond to the maximum and minimum $x$ coordinates of the associated net, and $y_{\min}$ and $y_{\max}$ correspond to the maximum and minimum $y$-coordinate of the associated net. The RUDY score assigned to a location $(x, y)$ is computed by aggregating RUDY scores over all nets $e\in E$.

\subsection{Machine learning and EDA}
As previously mentioned, ML-based prediction has been explored for various early-stage tasks in EDA flows including routability, DRC, and IR drop prediction~\cite{mavirec, congestion, routenet}. For the purposes of this work, we focus on the congestion prediction framework proposed by \citet{congestion}. Notably,  \citet{congestion} use the Innovus global router to obtain ground truth congestion hotspots, while an $W \times H \times 3$ feature map $M$, comprised of RUDY scores~\cite{rudy}, an associated pin-density variant PinRudy, and a macro placement map MacroRegion is derived from the associated cell placement. A neural network is used to learn a mapping from feature maps to congestion hotspots. The authors of~\cite{congestion} also derive the gradients of the unsupervised congestion penalty $\frac{1}{HW}||f_\theta(M)||_F^2$ with respect to cell locations.

\subsection{Machine learning and robustness}
\label{sec:ml_and_roubust}

Consider the network $f:\mathbb{R}^d \to \mathbb{R}^k$, where the input is $d$-dimensional and the output is a $k$-dimensional vector of likelihoods. For example, the input could be a $d$-dimensional image and the $j$-th entry of the output could correspond to the likelihood the image belongs to the $j$-th class. The associated prediction is then $c(x; \theta) = \arg\max_{j\in[k]} f_{j}(x; \theta)$. 

Recently, machine learning practitioners have not just been concerned that the prediction be correct, but also want robustness to random or adversarial noise, i.e. small perturbations to the input which may change the prediction to an incorrect class. We define the notion of $\epsilon$-robustness below:
\begin{definition}[$\epsilon$-robust]
$f$ parameterized by $\theta$ is called $\epsilon$-robust with respect to norm $p$ at $x$ if the prediction is consistent for a small ball of radius $\epsilon$ around $x$:
\begin{equation}
    \label{eq:erobust}
    c(x+\delta; \theta) = c(x; \theta),\quad \forall \delta : ||\delta||_p \leq \epsilon.
\end{equation}
\end{definition}
%
%
The minimal $\ell_p$-norm perturbation $\delta_p^*$ required to switch an sample's label is given by the solution to the following problem:
$$
\delta^*_p = \arg\min ||\delta||_p \quad \textrm{s.t.}\quad c(x; \theta) \neq c(x+\delta; \theta).
$$
A significant amount of existing work relies on a first-order approximations and H\"{o}lder's inequality to recover $\delta^*$.

Projected Gradient Descent (PGD) is a first-order method that can be used to find an approximation of $\delta^*_p$. PGD-type algorithms consist of a descent step followed by a projection onto the feasible set $S$. Given the current iterate $x^{(i)}$, the next iterate $x^{(i+1)}$ is computed via a transformation $s$ applied to the gradient of the loss function $L$. For example, if labels are available to the perturbation algorithm, $L$ could be the original loss used to train the network $f$. Alternatively, $L$ can be substituted for an unsupervised metric.
\begin{equation}
\begin{aligned}
    & u^{(i+1)} = x^{(i)} + \eta^{(i)}\cdot s(\nabla L(x^{(i)})) \\
    &x^{(i+1)} = P_{S}(u^{(i+1)})
\label{eq:pgd}
\end{aligned}
\end{equation}
$\eta^{(i)} > 0$ the step size at iteration $i$, $s : \mathbb{R}^d \to \mathbb{R}^d$ determines the descent direction as a function of the gradient of the loss $L$ at $x^{(i)}$ and $P_S : \mathbb{R}^d \to S$ is the projection on $S$. For example, an $\ell_1$-perturbation model of radius $\epsilon$, we denote by $B_\infty(x,\epsilon) = \{z \in \mathbb{R}^d |\:\: ||z - x||_\infty \leq \epsilon\}$.
A crucial choice is that of the descent direction $s(\nabla L(x_i))$, a mapping $s$ applied to a gradient. For example, the steepest descent direction~\cite{BoydVandenberghe}:
\begin{equation}
    \delta^*_{p} = \arg\max_{\delta \in \mathbb{R}^d}\langle w, \delta \rangle \quad \text{s.t.}\quad ||\delta||_p\leq \epsilon
\end{equation}
with $w  = \nabla f(x_i) \in \mathbb{R}^d $, the maximizer of a linear function over a given $\ell_p$ ball constraint. Thus one gets $\delta^*_\infty = \epsilon \text{sign}(w)$ and $\delta^*_2 = \epsilon w/ ||w||_2$ for $p = \infty$ and $p = 2$ respectively, which define $s$.

%
%

\section{A novel notion of imperceptibility for congestion prediction}
\label{sec:method}
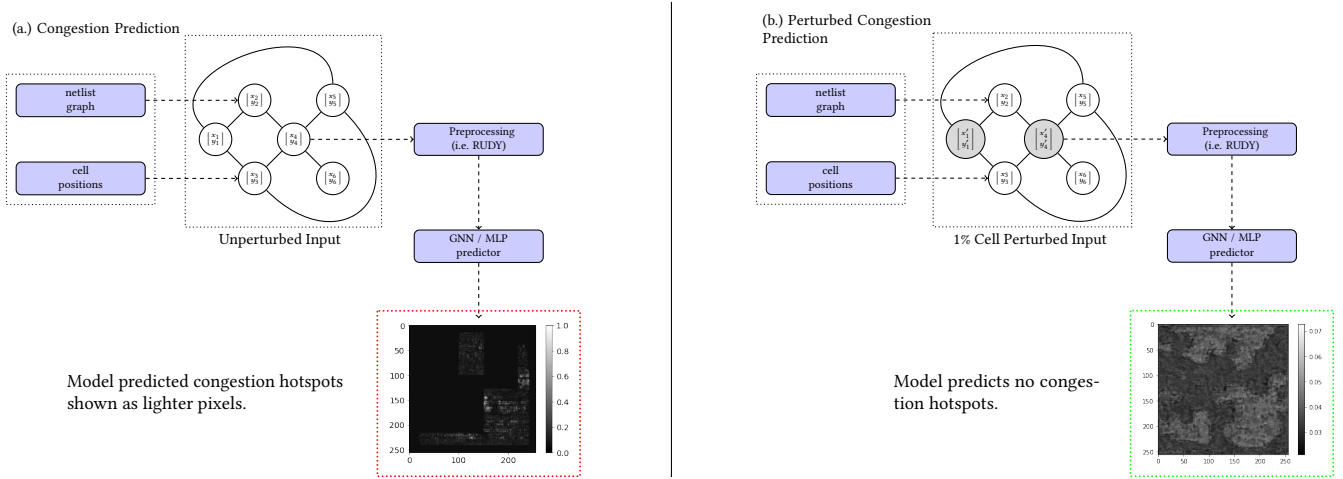
\begin{figure*}
    \centering
\scalebox{0.49}{
    \begin{tikzpicture}[node distance={15mm}, thick, main/.style = {draw, circle}] 
        \node[text width=5cm, font=\LARGE] at (-3,3) {(a.) Congestion Prediction};
        
        \node[main] (1) {\tiny$\left[\begin{smallmatrix}x_1\\y_1\end{smallmatrix}\right]$}; 
        \node[main] (2) [above right of=1] {\tiny$\left[\begin{smallmatrix}x_2\\y_2\end{smallmatrix}\right]$}; 
        \node[main] (3) [below right of=1] {\tiny$\left[\begin{smallmatrix}x_3\\y_3\end{smallmatrix}\right]$}; 
        \node[main] (4) [above right of=3] {\tiny$\left[\begin{smallmatrix}x_4\\y_4\end{smallmatrix}\right]$}; 
        \node[main] (5) [above right of=4] {\tiny$\left[\begin{smallmatrix}x_5\\y_5\end{smallmatrix}\right]$}; 
        \node[main] (6) [below right of=4] {\tiny$\left[\begin{smallmatrix}x_6\\y_6\end{smallmatrix}\right]$}; 
        
        \draw (1) -- (2); 
        \draw (1) -- (3); 
        \draw (1) to [out=135,in=90,looseness=1.5] (5); 
        \draw (2) -- (4); 
        \draw (3) -- (4); 
        \draw (5) -- (4); 
        \draw (5) to [out=315, in=315, looseness=2.5] (3); 
        \draw (6) -- node[midway, above right, sloped, pos=1] {} (4); 
        
        \draw[thick,dotted]     ($(1.north west)+(-0.5,2.5)$) rectangle ($(6.south east)+(1.0, -1.0)$);
        
        \node[block, left = 2.5cm of 2]   (i1){netlist graph};
        \node[block, left = 2.5cm of 3]   (i2){cell positions};
        \draw[thick,dotted]     ($(i1.north west)+(-0.25,0.25)$) rectangle ($(i2.south east)+(0.25, -0.25)$);
        \draw[->, dashed] (i1) -- (2); 
        \draw[->, dashed] (i2) -- (3); 
        
        \node[block, right = 2.8cm of 4]   (rudy){Preprocessing (i.e. RUDY)};
        \node[block, below = 2cm of rudy]   (gnn){GNN / MLP predictor};
        \draw[->, dashed] (4) -- (rudy); 
        \draw[->, dashed] (rudy) -- (gnn); 
        \node[text width=5cm, align=center] at (1.75,-2.75) (cap_graph){\LARGE Unperturbed Input};
        
        \node[inner sep=0pt, below = 1.5cm of gnn] (cong)
    {\includegraphics[width=.285\textwidth]{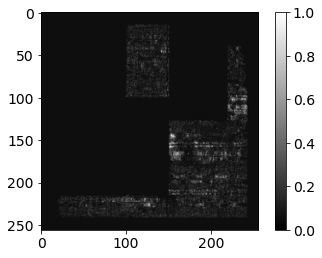}};
        \node[left = 1cm of cong, text width=7.5cm, align=justify, font=\huge]   (cap_cong){Model predicted congestion hotspots shown as lighter pixels.};
        \draw[->, dashed] (gnn) -- (cong); 
        \draw[very thick,dotted, color=red]     ($(cong.north west)+(-0.2,0.2)$) rectangle ($(cong.south east)+(0.2, -0.2)$);
    \end{tikzpicture} 
}
\hspace{1cm} \vrule \hspace{1cm}
\scalebox{0.49}{
    \begin{tikzpicture}[node distance={15mm}, thick, main/.style = {draw, circle}] 
        \node[text width=5cm, , font=\LARGE] at (-3,3) {(b.) Perturbed Congestion Prediction};
        
        \node[main, fill=gray!30] (1) {\tiny$\left[\begin{smallmatrix}x_1'\\y_1'\end{smallmatrix}\right]$}; 
        \node[main] (2) [above right of=1] {\tiny$\left[\begin{smallmatrix}x_2\\y_2\end{smallmatrix}\right]$}; 
        \node[main] (3) [below right of=1] {\tiny$\left[\begin{smallmatrix}x_3\\y_3\end{smallmatrix}\right]$}; 
        \node[main, fill=gray!30] (4) [above right of=3] {\tiny$\left[\begin{smallmatrix}x_4'\\y_4'\end{smallmatrix}\right]$}; 
        \node[main] (5) [above right of=4] {\tiny$\left[\begin{smallmatrix}x_5\\y_5\end{smallmatrix}\right]$}; 
        \node[main] (6) [below right of=4] {\tiny$\left[\begin{smallmatrix}x_6\\y_6\end{smallmatrix}\right]$}; 
        
        \draw (1) -- (2); 
        \draw (1) -- (3); 
        \draw (1) to [out=135,in=90,looseness=1.5] (5); 
        \draw (2) -- (4); 
        \draw (3) -- (4); 
        \draw (5) -- (4); 
        \draw (5) to [out=315, in=315, looseness=2.5] (3); 
        \draw (6) -- node[midway, above right, sloped, pos=1] {} (4); 
        
        \draw[thick,dotted]     ($(1.north west)+(-0.5,2.5)$) rectangle ($(6.south east)+(1.0, -1.0)$);
        
        \node[block, left = 2.5cm of 2]   (i1){netlist graph};
        \node[block, left = 2.5cm of 3]   (i2){cell positions};
        \draw[thick,dotted]     ($(i1.north west)+(-0.25,0.25)$) rectangle ($(i2.south east)+(0.25, -0.25)$);
        \draw[->, dashed] (i1) -- (2); 
        \draw[->, dashed] (i2) -- (3); 
        
        \node[block, right = 2.8cm of 4]   (rudy){Preprocessing (i.e. RUDY)};
        \node[block, below = 2cm of rudy]   (gnn){GNN / MLP predictor};
        \draw[->, dashed] (4) -- (rudy); 
        \draw[->, dashed] (rudy) -- (gnn);
        \node[text width=5cm, align=center] at (1.75,-2.75) (cap_graph){\LARGE 1\% Cell Perturbed Input};
        
        \node[inner sep=0pt, below = 1.5cm of gnn] (cong)
    {\includegraphics[width=.285\textwidth]{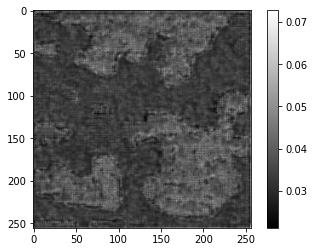}};
        \node[left = 1cm of cong, text width=5.5cm, align=justify, font=\huge]   (cap_cong){Model predicts no congestion hotspots.};
    
        \draw[->, dashed] (gnn) -- (cong); 
        \draw[very thick,dotted, color=green]     ($(cong.north west)+(-0.2,0.2)$) rectangle ($(cong.south east)+(0.2, -0.2)$);
        
    \end{tikzpicture}  
}

    \caption{General illustration of effect of imperceptible perturbations on EDA predictions \textbf{(a.)}: Vanilla prediction framework. The netlist-graph and cell attributes (i.e. positions) are used to make predictions (e.g. DRC locations or congestion hotspots) via the neural predictor. \textbf{(b.)}: The attributes of a subset of nodes are perturbed: $x' = x + \delta_x$. The predictor is vulnerable (i.e. can be made to predict that a design is congestion-free)\textemdash even when both $\delta_x$ and the number of perturbed nodes are small.}
    \label{fig:graph}
\end{figure*}
In this section, we describe a method to compute layout perturbations that guarantee consistency of the global routing. We present a general illustration of our method in Fig.~\ref{fig:graph}. In other words, given a trained model, we seek to adjust a given layout (the coordinates of a subset of the cells in a design) such that the adjusted layout remains valid, but spoils the congestion predictions of the model.

%
\begin{figure}
    \centering
    \includegraphics[width=0.4\textwidth]{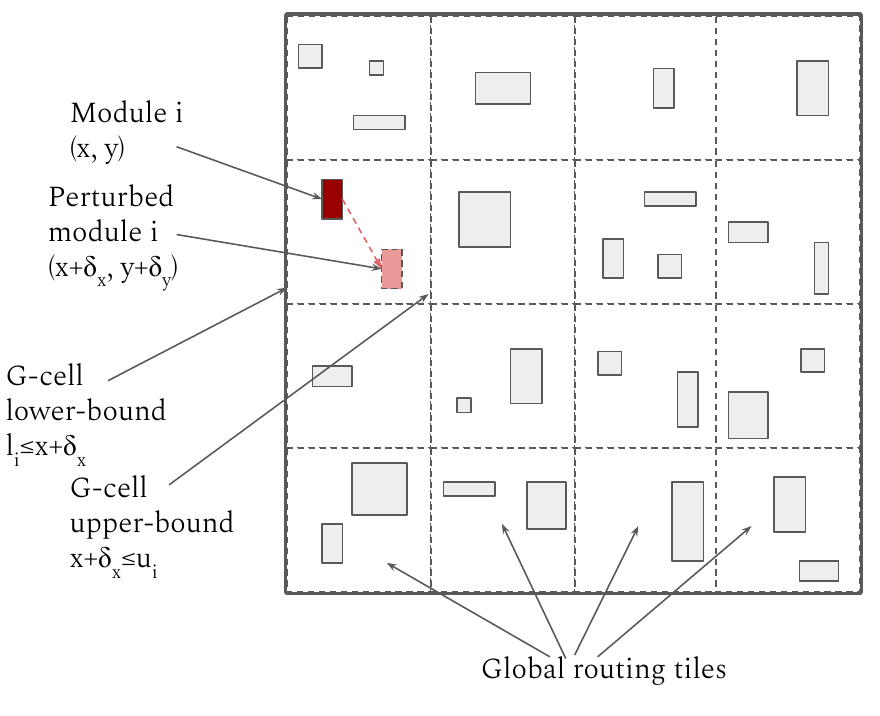}
    \caption{Local constraints for each movable cell (highlighted in red) ensures cells do not move G-Cells.}
    \label{fig:constraints}
    \vspace{-0.4cm}
\end{figure}
%
In particular, we define the feasible set of perturbations that we utilize in the context of congestion predictions and an algorithm to perform a projection onto this set. \emph{Specifically, we discuss perturbations of coordinate-based representations of chip layouts}. Inspired by earlier work on adversarial attacks designed for image classifiers~\cite{croce2020reliable,croce2021mind,croce2019sparse}, we impose a natural set of constraints on the perturbation to ensure that the global routing does not change. Individually, these constraints prohibit cells from moving outside of their G-Cell tiles.

Let $X := [x : y]^\top \in \mathbb{R}^{n\times 2}$ be the coordinate assignments to each cell and $\delta := [\delta_x : \delta_y]^\top \in \mathbb{R}^{n\times 2}$ be a perturbation. For simplicity, and without loss of generality, let us consider the 1st column of $X$ and $\delta$; $x, \delta_x$ (i.e. the $x$-coordinate of each cell and associated perturbations such that the perturbed cell $x$-coordinates of the layout is given by $x + \delta_x$). 

We say that $\delta_x$ is an \emph{imperceptible} perturbation in the context of physical design if the following conditions hold:

\begin{enumerate}
    \item Global imperceptibility: the maximum number of cells that can be moved is bounded by $\epsilon_0$: $||\delta||_0 \leq \epsilon_0$ (e.g. $1\%$ of the total number of cells.
    \item Local imperceptibility: a cell can only be moved within it's associated G-Cell: $l_i \leq (x + \delta_x)_i \leq u_i$
\end{enumerate}
%

Given $x \in [0,1]^d$, we define the feasible set\textemdash the intersection of the $\ell_0$-ball of radius $\epsilon_0$, the upper and lower-bound\textemdash \emph{box}\textemdash constraints on each entry of $x$, and the set $[0,1]^d$:
\begin{equation}
\begin{aligned}
S(x) = \Bigl\{&z \in \mathbb{R}^d | \:\: \sum_{i=1}^d \mathbf{1}_{|z_i - x_i| > 0}\leq \epsilon_0, \\
&l_i \leq z_i - x_i \leq u_i,\:\:0 \leq z_i \leq 1 \Bigl\}
\end{aligned}
\end{equation}
%
%
First, note that intersection of the two constraints $0 \leq z_i \leq 1$ and $l_i \leq z_i - x_i \leq u_i$ may be written:
$$
\max\{0, l_i + x_i\} \leq z_i \leq \min\{1,u_i + x_i\} 
$$
So, the feasible set may be simplified and re-written
\begin{equation}
\begin{aligned}
S(x) = \Bigl\{&z \in \mathbb{R}^d | \sum_{i=1}^d \mathbf{1}_{|z_i - x_i| > 0}\leq \epsilon_0, \\
&\max\{0, l_i + x_i\} \leq z_i \leq \min\{1, u_i + x_i\} \Bigl\}
\end{aligned}
\end{equation}
The Euclidean projection onto $S(x)$; $P_S$  is then defined to be
\begin{equation}
\begin{aligned}
\min_z ||y - z||_2^2\quad
&\text{s.t. } \sum_{i=1}^d \mathbf{1}_{|z_i - x_i| > 0}\leq \epsilon_0, \\
&\max\{0, l_i + x_i\} \leq z_i \leq \min\{1, u_i + x_i\}
\end{aligned}\label{eq:Sproj}
\end{equation}
Ignoring the combinatorial constraint, the solution is given by $z_{i}^* = \max\{l_i + x_i, \min\{y_i, u_i + x_i\}\}$. We re-integrate the $\ell_0$ constraint and resolve the projection by sorting according to the \emph{gain} $\phi$:
$$
\phi_i = (y_i - x_i)^2 - (y_i - z_i^*)^2,\quad
z_{\pi_i} = 
\begin{cases}
z^*_{\pi_i} \quad i=1,\ldots,k \\
x_{\pi_i} \quad \text{otherwise}
\end{cases}
$$
Thus, the final solution will have $k$ entries\textemdash those with the highest gain\textemdash and that differ by no more than $l_i$ or $u_i$. 

Importantly, the solution to this problem can be computed efficiently; requiring a single backward pass through the trained model to compute the perturbed layout $y$ and computation of the projection $P_S$ in linear time\textemdash note that when ignoring the combinatorial constraint, Prob.~\ref{eq:Sproj} is a separable problem (can be computed in parallel) over the cells. Finding the subset of cells to shift (satisfying the $\ell_0$ constraint) involves a linear-time scan over gains.

\section{Experiments}
\label{sec:experiments}
\begin{table}
\caption{Vulnerability of ML-based EDA predictors to imperceptible perturbations. ``*'' denotes a perturbation that induces congestion-free predictions. ``$\dag$'' denotes a perturbation that induces mispredictions of hotspots.}
\label{tab:results}
\centering
\resizebox{0.49\textwidth}{!}{%
\begin{tabular}{|l| l l l| l l l|} 
\hline

\cline{2-7}

     & $\frac{1}{HW}||M||^2_F$    & NRMS    & SSIM &  $\frac{1}{HW}||M||^2_F$    & NRMS    & SSIM     \\
\hline

\textit{FCN Model} & \multicolumn{3}{c|}{Vanilla training} & \multicolumn{3}{c|}{Robust training}         \\ 

\hline
Vanilla / none   & 0.010  & 0.0393  & 0.8044  &  0.010  & 0.0393  & 0.7970 \\
Random noise    &   0.012  & 0.0420  & 0.7970  & 0.012   &  0.0420 & 0.7123 \\
* 1\% cells perturbed    & 0.0012   &  0.0791   & 0.6255  &  0.011  & 0.0561  & 0.6831 \\
* 5\% cells perturbed    & 0.0011   &  0.0945   & 0.5152  & 0.012  & 0.0533  & 0.6181 \\
\dag 1\% cells perturbed     &  0.011  &  0.1055   &  0.4534  & 0.011  &  0.0431 & 0.7011 \\
\dag 5\% cells perturbed     & 0.011   &  0.1467   &  0.4334  &  0.011  & 0.0440 & 0.7193 \\
\hline


\textit{GNN Model}      &    &    &  &     &     &      \\
\hline
Vanilla / none   &   0.011  & 0.0348  & 0.8130  & 0.011   & 0.0393  & 0.7970 \\
Random noise    &  0.013  & 0.0384  & 0.7974  &  0.011  &  0.0417 & 0.6933 \\
* 1\% cells perturbed    & 0.0013   &  0.0743   & 0.6129  & 0.0098  & 0.0403  & 0.7643 \\
* 5\% cells perturbed    & 0.0012   &  0.0892   & 0.5032  & 0.0097  & 0.0407 & 0.7392 \\
\dag 1\% cells perturbed     &  0.011  &  0.1744   &  0.4461  &  0.011 & 0.0411  & 0.694  \\
\dag 5\% cells perturbed     & 0.013   &  0.1835   &  0.4219  & 0.013  & 0.0417  & 0.695 \\
\hline

\end{tabular}
}
\end{table}
\begin{table}[htb]
\small
\centering
\caption{From~\cite{chai2022circuitnet}. Statistics of designs and variations.}
\resizebox{0.49\textwidth}{!}{%
\begin{tabular}{|cccc||cc|}
\hline
\multicolumn{1}{|c|}{\multirow{2}{*}{Design}} & \multicolumn{3}{c||}{Netlist Statistics} & \multicolumn{2}{c|}{Synthesis Variations} \\ \cline{2-6}
\multicolumn{1}{|c|}{} & \multicolumn{1}{c|}{\#Cells} & \multicolumn{1}{c|}{\#Nets} & \multicolumn{1}{c||}{\begin{tabular}[c]{@{}c@{}}Cell Area\\ ($\mu m^2$)\end{tabular}} & \multicolumn{1}{c|}{\#Macros} & \begin{tabular}[c]{@{}c@{}}Frequency\\ (MHz)\end{tabular} \\ \hline
\multicolumn{1}{|c|}{RISCY-a} & \multicolumn{1}{c|}{44836} & \multicolumn{1}{c|}{80287} & \multicolumn{1}{c||}{65739} & \multicolumn{1}{c|}{\multirow{3}{*}{3/4/5}} & \multirow{6}{*}{50/200/500} \\ \cline{1-1}
\multicolumn{1}{|c|}{RISCY-FPU-a} & \multicolumn{1}{c|}{61677} & \multicolumn{1}{c|}{106429} & \multicolumn{1}{c||}{75985} & \multicolumn{1}{c|}{} &  \\ \cline{1-1}
\multicolumn{1}{|c|}{zero-riscy-a} & \multicolumn{1}{c|}{35017} & \multicolumn{1}{c|}{67472} & \multicolumn{1}{c||}{58631} & \multicolumn{1}{c|}{} &  \\ \cline{1-1} \cline{5-5}
\multicolumn{1}{|c|}{RISCY-b} & \multicolumn{1}{c|}{30207} & \multicolumn{1}{c|}{58452} & \multicolumn{1}{c||}{69779} & \multicolumn{1}{c|}{\multirow{3}{*}{13/14/15}} &  \\ \cline{1-1}
\multicolumn{1}{|c|}{RISCY-FPU-b} & \multicolumn{1}{c|}{47130} & \multicolumn{1}{c|}{84676} & \multicolumn{1}{c||}{80030} & \multicolumn{1}{c|}{} &  \\ \cline{1-1}
\multicolumn{1}{|c|}{zero-riscy-b} & \multicolumn{1}{c|}{20350} & \multicolumn{1}{c|}{45599} & \multicolumn{1}{c||}{62648} & \multicolumn{1}{c|}{} &  \\ \hline \hline
\multicolumn{6}{|c|}{Physical Design Variations} \\ \hline
\multicolumn{1}{|c|}{\begin{tabular}[c]{@{}c@{}}Utilizations\\ (\%)\end{tabular}} & \multicolumn{2}{c|}{\begin{tabular}[c]{@{}c@{}}\#Macro\\ Placement\end{tabular}} & \multicolumn{1}{c|}{\begin{tabular}[c]{@{}c@{}}\#Power Mesh\\ Setting\end{tabular}} & \multicolumn{2}{c|}{Filler Insertion} \\ \hline
\multicolumn{1}{|c|}{70/75/80/85/90} & \multicolumn{2}{c|}{3} & \multicolumn{1}{c|}{8} & \multicolumn{2}{c|}{\begin{tabular}[c]{@{}c@{}}After Placement\\ /After Routing\end{tabular}} \\ \hline
\end{tabular}
}
\label{tab:datasets}
\end{table}
In this section we describe a set of comprehensive experiments on  testcases from the CircuitNet suite~\cite{chai2022circuitnet}. Summary statistics of the testcases are presented in Table~\ref{tab:datasets}. Our numerical experiments are aimed at establishing the efficacy of our method with respect to spoiling congestion predictions made by two trained architectures.

\subsection{Experimental setup}

We evaluate the impact of our small perturbations on the robustness of ML-based congestion predictors. Moreover, we give illustrative examples of such sparse and imperceivable perturbations. We utilize the CircuitNet benchmarks~\cite{chai2022circuitnet} to validate our method. CircuitNet is an open-source dataset consisting of more than $10$K samples from versatile runs of commercial design tools based on open-source RISC-V designs with various features for multiple ML for EDA applications. We summarize the designs and generation procedure used to compose the CircuitNet dataset in Table~\ref{tab:datasets}.

Perturbations are generated using a momentum-based PGD algorithm with restarts. We adapt the standard PGD iterations outlined in Eq.~\ref{eq:pgd} in two ways: (1.) we adjust the gradient-based update rule to incorporate a momentum term:
\begin{equation}
\begin{aligned}
& u^{(i+1)} = P_{S}(x^{(i)} + \eta^{(i)}\cdot s(\nabla L(x^{(i)}))) \\
&x^{(i)} = P_{S}(x^{(i)} + \alpha(u^{(i+1)} - x^{(i)}) + (1-\alpha)(x^{(i)} - x^{(i-1)}))
\end{aligned}
\end{equation}
where $\alpha\in [0,1]$ regulates the influence of the previous update on the current one and $P_S$ is described in Sec.~\ref{sec:method}. (2.) we introduce ``restarts''\textemdash i.e. we apply PGD to several random initializations and select the best solution.

\subsubsection{Algorithm parameters} We adopt the same $70/30$ train-test split described in the CircuitNet paper. We perturb all samples in the test. In Table~\ref{tab:results} we report several metrics for each method including our congestion score, NRMS, and SSIM. We set $\alpha = 0.75$ and fix $\eta$ to be $2\cdot w$, where $w$ is the width of each G-Cell. Each perturbed example is generated by running PGD for $100$ iterations with $5$ restarts, and $\eta$ is linearly decayed to $1/10 \cdot w$.


\subsection{On the robustness of congestion predictors}
\begin{figure}
    \centering
    \begin{subfigure}[b]{0.495\linewidth}
    \includegraphics[width=\linewidth]{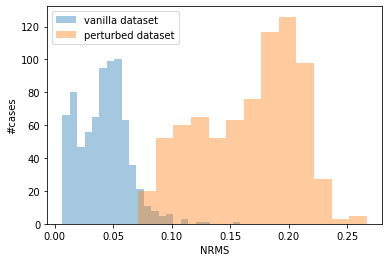}
    \subcaption{}
    \label{fig:nrms}
    \end{subfigure}
    \begin{subfigure}[b]{0.495\linewidth}
    \includegraphics[width=\linewidth]{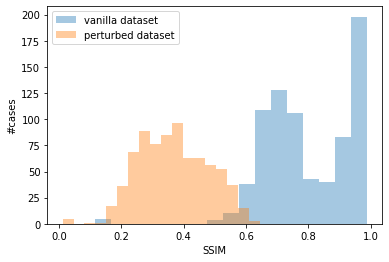}
    \subcaption{}
    \label{fig:ssim}
    \end{subfigure}
    \caption{Performance metrics associated with a vanilla network evaluated on unperturbed and perturbed layouts. (\textbf{\subref{fig:nrms}}) Distribution shift in NRMS. (\textbf{\subref{fig:ssim}}) Distribution shift in SSIM. \citet{congestion} characterize a good predictor as achieving $NRMS < 0.2$ and $SSIM > 0.8$. Using our method, we are able create valid inputs such that approximately $100\%$ of samples have $SSIM < 0.8$ and $60\%$ of samples have $NRMS > 0.15$. $100\%$ of samples satisfy one of the two conditions.}
    \label{fig:scoredists}
\end{figure}
\begin{figure}
    \centering
    \includegraphics[width=0.4\textwidth]{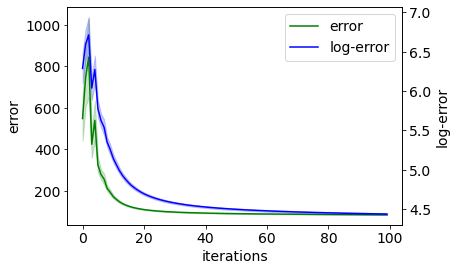}
    \caption{Mean relative unsupervised error of PGD over iterations. Shaded region denotes 1 standard deviation. Note the log-scale in blue implies convergence in error.}
    \label{fig:my_label}
    \vspace{-0.4cm}
\end{figure}
In Table~\ref{tab:results}, we evaluate our method for producing valid and imperceptible perturbations using the fully convolutional architecture proposed in~\citet{congestion} and a single-layer graph convolutional network (GNN Model) proposed in~\cite{kipf2017semisupervised}. Two variants of our algorithm are implemented. For rows denoted by a *, the perturbation ascent direction is computed with respect to the congestion score $\frac{1}{HW}||f_\theta(M)||_F^2$. For rows denoted by a $\dagger$, the perturbation ascent direction is computed with respect to the \emph{supervised prediction loss}. 

We demonstrate that neural network predictors fail to accurately predict congestion of layouts produced by our method. Furthermore, when a larger budget of cells are allowed to be shifted, performance is further degraded. In particular, we first observe that vanilla models are relatively robust to a random perturbation model. Namely, we uniformly at random select 1\% of cells and maximally perturb their associated positions such that they remain within their associated G-Cell. When the layout is altered in this way, we see that neural predictors generally maintain their performance with only a minor degradation in NRMS and SSIM observed. 

Next, we demonstrate that perturbations may be carefully chosen such that the associated predictions correspond to congestion-free predictions, or even adversarial predictions- i.e. the model predicts congestion in regions which are congestion-free and predicts congestion-free regions in areas which are highly congested (e.g. in regions with macros). We provide examples to demonstrate these instances in Figure~\ref{fig:layout}. Distributions of SSIM and NRMS scores are provided in Figure~\ref{fig:scoredists}. Notably, the predictor violates the conditions necessary for good performance (NRMS $< 0.2$, SSIM $> 0.8$)~\cite{chai2022circuitnet,fpgacongestion}. 

When a budget of $1\%$ of cells is prescribed, a degradation in SSIM of $43.64\%$ and a degradation in NRMS of $168.45\%$ are observed for the FCNN-based predictor. Likewise, when the budget is increased to $5\%$ of cells, degradations in SSIM and NRMS amount to $46.12\%$ and $273.28\%$ are observed respectively. As expected, the GNN-based model is also vulnerable to the aforementioned issues. Interestingly, while the GNN-based method outperforms the FCN-based method on unperturbed layouts, the GNN-based model is seemingly \emph{more} vulnerable to our proposed method with degradations in NRMS and SSIM of up to $427.3\%$ and $48.11\%$ for a budget of $5\%$ of cells. 

\subsection{Improving robustness of congestion predictors via momentum-based PGD}
\begin{figure}
    \centering
    \includegraphics[width=0.4\textwidth]{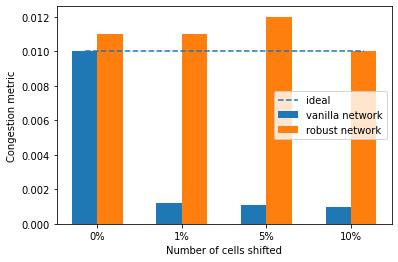}
    \caption{Robust congestion prediction statistic ($\frac{1}{HW}||f_\theta(M)||_F^2$) and percentage of cells that are allowed to move (looseness of $\epsilon_0$ constraint).}
    \label{fig:scorehists}
    \vspace{-0.5cm}
\end{figure}
\begin{figure}[ht]
\centering
\begin{subfigure}[b]{0.54\linewidth}
\includegraphics[width=\linewidth]{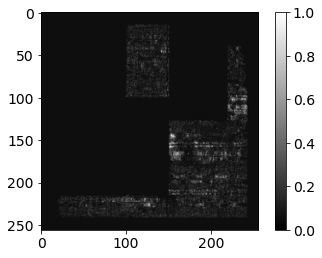}
\subcaption{}
\label{fig:good1}
\end{subfigure}
\begin{subfigure}[b]{0.45\linewidth}
\includegraphics[width=\linewidth]{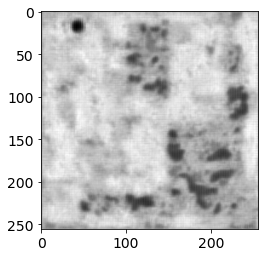}
\subcaption{}
\label{fig:bad1}
\end{subfigure}

\vspace{.1cm}

\begin{subfigure}[b]{0.49\linewidth}
\includegraphics[width=1.1\linewidth]{figures/example_good2.png}
\subcaption{}
\label{fig:good2}
\end{subfigure}
\hfill
\begin{subfigure}[b]{0.48\linewidth}
\includegraphics[width=1.1\linewidth]{figures/example_bad2_2.png}
\subcaption{}
\label{fig:bad2}
\end{subfigure}

\caption{(\textbf{\subref{fig:good1}\textemdash\subref{fig:bad1}}): Predictions for robust and nonrobust estimators (supervised loss). Both plots share the same gradient scale. (\textbf{\subref{fig:good2}\textemdash\subref{fig:bad2}}) Predictions for robust and nonrobust estimators (unsupervised loss). Note the difference in gradient scale.}
\label{fig:layout}
\vspace{-0.5cm}
\end{figure}
A number of techniques~\cite{madry2018towards,carliniwagner17,Wang2020Improving} have been proposed to mitigate the issue of robustness of deep networks, with some of the most reliable being certified defenses~\cite{CohenSmooth3} and methods based on the principle of adversarial training~\cite{madry2018towards}. In this work, we forgo investigating provable defenses and instead stick with PGD-based adversarial training. More concretely, each iteration of SGD, a portion of each batch (i.e. $50\%$) is perturbed via our method. Running PGD during training is expensive. One may exploit the Fast-FGSM algorithm proposed in~\cite{Wang2020Improving}, which demonstrates that by simply introducing random initialization points from which to compute adversarial perturbations, one projected gradient step is as effective as repeated steps during training while being significantly more efficient.

Using standard PGD, we train models on using the CircuitNet split and report the \emph{robust test statistics} in the left column of Table~\ref{tab:results}. We see that models trained using adversarial training are \emph{significantly} more robust with respect to both unsupervised and supervised perturbations. More concretely, we observe recovery of predictive quality primarily with respect to the metric driving perturbations (i.e. when predictors are trained to be robust to the unsupervised congestion metric, robustness is improved across all metrics, but most significantly for unsupervised congestion). In Figure~\ref{fig:scorehists}, we provide a comparison between a vanilla network and a robust network trained via PGD. On the $x$-axis, we plot the $\epsilon_0$ constraint, the percentage of cells that are free to be adjusted. On the $y$-axis, we plot the mean congestion metric $\frac{1}{HW}||f_\theta(M)||_F^2$ across the validation set. Note that the robust network maintains good performance, even as the number of cells increases.

The congestion value $0.01$ is from a baseline predictor\textemdash the blue bar at 0\% cells shifted. Note that the value may not be representative of prediction quality (instead, see Figure~4). Ideally bars should be the same height across perturbations. However, the vanilla model's predictions on perturbed layouts change significantly. In contrast, robust predictors generalize (orange bars have similar height).

\section{Conclusion and Future Work}
\label{sec:conclusion}

In this paper, we have demonstrated that CNN-based congestion detection models are vulnerable to small perturbations. We have proposed and evaluated layout perturbations that are \emph{guaranteed} to not alter a early global routing. To address these issues, we have proposed to apply adversarial training, demonstrating that such methods can improve robustness and generalization of deep learning-based EDA systems. The implication of our work is that designers should carefully evaluate deep learning-based models when employing ML-based CAD systems in EDA pipelines.
More broadly, we hope that our perturbation methodology for evaluating vulnerabilities in congestion prediction and our adversarial training approach to make congestion predictions more robust can be adapted for evaluating ML-based EDA predictors. 

\begin{acks}
We acknowledge support from NSF CCF-2110419. 
\end{acks}
\bibliographystyle{ACM-Reference-Format}
\bibliography{main}
\appendix

\end{document}